\PassOptionsToPackage{dvipsnames}{xcolor}
\documentclass[10pt, a4paper]{article}
\usepackage{lrec-coling2024} 

\usepackage{natbib}
\usepackage{multibib}
\makeatletter
\def\@mb@citenamelist{cite,citep,citet,citealp,citealt,citepalias,citetalias}
\makeatother
\newcites{languageresource}{~}

\usepackage{graphicx}
\usepackage{tabularx}
\usepackage{soul}

\usepackage{hyperref}

\usepackage{amsmath}

\usepackage[inline,shortlabels]{enumitem}
\usepackage[normalem]{ulem}

\usepackage{multirow}
 \usepackage{amssymb}

\newcommand{\gradientcell}[6]{
    \ifdimcomp{#1pt}{>}{#3 pt}{\cellcolor{#5!100.0!#4!#6}#1}{
    \ifdimcomp{#1pt}{<}{#2 pt}{\cellcolor{#5!0.0!#4!#6}#1}{
         \pgfmathparse{int(round(100*(#1/(#3-#2))-(#2 *(100/(#3-#2)))))}
        \xdef\tempa{\pgfmathresult}
        \cellcolor{#5!\tempa!#4!#6}\hspace*{-3.5pt}#1\hspace*{-4.5pt}
    }}
 }

\usepackage{pgf} 
\usepackage{colortbl}

\newcommand{\Rp}[1]{\gradientcell{#1}{.00}{.99}{Cyan}{Yellow}{70}}
\newcommand{\Av}[1]{\gradientcell{#1}{1.9}{4.1}{Cyan}{Yellow}{70}}
\newcommand{\St}[1]{\gradientcell{#1}{.4}{2}{Cyan}{Yellow}{70}}

\usepackage{xcolor}
\usepackage{hyperref}
 \definecolor{darkblue}{rgb}{0, 0, 0.5}
  \hypersetup{colorlinks=true, citecolor=darkblue, linkcolor=darkblue, urlcolor=darkblue}

\usepackage{xstring}

\usepackage{color}
\definecolor{green-CultLib}{HTML}{008000}

\title{How Gender Interacts with Political Values: \\ A Case Study on Czech BERT Models}

\name{Adnan Al Ali and Jindřich Libovický} 

\address{Charles University, Faculty of Mathematics and Physics, Institute of Formal and Applied Linguistics \\
         Malostranské náměstí 25, 118 00 Praha, Czech Republic\\
         adnan@matfyz.cz \quad libovicky@ufal.mff.cuni.cz \\
         }

\abstract{
Neural language models, which reach state-of-the-art results on most natural language processing tasks, are trained on large text corpora that inevitably contain value-burdened content and often capture undesirable biases, which the models reflect.
This case study focuses on the political biases of pre-trained encoders in Czech and compares them with a representative value survey. Because Czech is a gendered language, we also measure how the grammatical gender coincides with responses to men and women in the survey.
We introduce a novel method for measuring the model's perceived political values. We find that the models do not assign statement probability following value-driven reasoning, and there is no systematic difference between feminine and masculine sentences. 
We conclude that BERT-sized models do not manifest systematic alignment with political values and that the biases observed in the models are rather due to superficial imitation of training data patterns than systematic value beliefs encoded in the models.
 \\ \newline  }

\begin{document}

\maketitleabstract

\section{Introduction}

The increasing use of pre-trained language models (LMs) leads to questions of how the models align with human values \citep{long2022instruct}.
LMs were found to exhibit certain biases, especially concerning gender and ethnicity \citep{stereoset}. Moreover, there are convincing arguments that the texts used as training data necessarily reflect the values and culture of the text authors; the model will likely reproduce these values \citep{stochastic-parrots}. 
%
In the real world, biases against groups of people are often a consequence of individuals' political and moral beliefs.
Whether the gender bias observed in LMs is due to systematic alignment with corresponding political values or a just result of mimicking surface patterns from training data is, however, unclear.

This paper contributes to this discussion with a case study that links the value alignment of pre-trained LMs with gender bias. We work with masked LMs trained for Czech. Czech is a particularly interesting language for this study because it is relatively high-resourced (with multiple RoBERTa-sized models available) and is strongly gendered: it has gendered nouns with agreement in gender with verbs and adjectives. Moreover, multilingual encoders have been shown to generate biased and toxic completions in Czech \citep{martinkova-etal-2023-measuring}.

We experiment with masked language models for Czech, all in their \emph{base} version: 
\begin{enumerate*}[a)]
    \item monolingual -- \emph{RobeCzech} \citep{robeczech}, \emph{Czert} \citep{czert}, \emph{FERNET News} \citep{fernet};
    \item multilingual -- \emph{Multilingual BERT} \citep{bert}, \emph{XLM-RoBERTa} \citep{xlm-r}, \emph{Slavic BERT} \citep{slavic-bert}
\end{enumerate*}. We evaluate to what extent the probability they assign to value judgments aligns with data from a representative survey \citep{toxoplasmosis} and how those get affected by the gender of the assumed author of the claims. Our results show that: 
\begin{enumerate*}[(1)]
    \item The models do not make a significant difference between the genders of the assumed author.
    \item The models' ratings of the statements corresponding to the same political value have a large variance, suggesting random behavior rather than presumed systematic political beliefs.
\end{enumerate*}

\section{Related Work}

LMs and models derived from them are known to contain many biases, although the definitions and conceptualizations of biases differ \citep{blodgett-etal-2020-language}.

Gender bias is probably the most intensively studied bias in NLP for several years in many tasks, including named entity recognition \citep{zhao-etal-2018-gender}, machine translation \citep{stanovsky-etal-2019-evaluating} or image captioning \citep{zhao-etal-2017-men}. 

The bias propagates to the models already from pre-trained representations, such as static \citep{bolukbasi2016man} or contextual embeddings \citep{zhao-etal-2019-gender}. There are several attempts to quantify the bias better: \citet{stereoset} and \citet{nangia-etal-2020-crows} present datasets for measuring stereotypical biases in various categories, including gender, and show that contextual embeddings exhibit strong stereotypical biases. \citet{martinkova-etal-2023-measuring} inspect gender bias in Czech, Polish, and Slovak language models and find that they produce more hurtful content regarding men, who are more often associated with death, violence, and sickness by the models.
Unlike most previous work, we go beyond documenting the bias and search for connections with political values encoded in the models.

LMs are also known to have political biases. \citet{feng-etal-2023-pretraining} used the Political Compass to assess the political leaning of LMs, concluding that most LMs are in the middle of the left-right access and differ in how libertarian and authoritarian they are. Other studies approach values in LMs in general. \citet{schramowski22language,haemmerl-etal-2023-speaking} study moral values in sentence embeddings, showing a high degree of alignment with everyday moral intuitions. \citet{arora-etal-2023-probing} focus on cultural differences in everyday moral judgments, showing some level of alignment in general; however, without adequately capturing differences between countries. 

Although there are convincing arguments that LMs reflect moral and political values from the training data \citep{stochastic-parrots} and several studies show general trends supporting it \citep{feng-etal-2023-pretraining,schramowski22language}, there is little evidence for  correlation or even causal relation of the observed biases with the values in the models, which we study in this paper.

\section{Methodology}

Our methodology compares the model predictions with data from a representative sociological survey \citep{toxoplasmosis}, asking political value questions, which are then aggregated into four major categories. We do so by prompting the models to agree and disagree with opinion statements and comparing their probabilities. The probabilities from the masked LMs cannot be used directly because there is a strong correlation between agreeing and disagreeing with a statement. We conduct a series of steps to eliminate this correlation. This makes our method more robust than in previous work \citep{stereoset,arora-etal-2023-probing}, which directly work with uncalibrated log-probabilities.

\subsection{Prompting the Models} \label{political-experiment-setup}
\begin{table*}[ht]
    \centering
    
    \begin{tabular}{ c | l l }
     & \textbf{Fem} & \textbf{Masc} \\
     \hline
    \textbf{ Agree} & [CS] \underline{Řekla}, že \uwave{souhlasí} s tím, že \_\_\_ & [CS] \underline{Řekl}, že \uwave{souhlasí} s tím, že \_\_\_ \\
     & \textit{\underline{She said} that she \uwave{agrees} that} \_\_\_ & \textit{\underline{He said} that he \uwave{agrees} that} \_\_\_ \\\hline
     \textbf{Disagree} & [CS] \underline{Řekla}, že \uwave{nesouhlasí} s tím, že \_\_\_ & [CS] \underline{Řekl}, že \uwave{nesouhlasí} s tím, že \_\_\_ \\
     & \textit{\underline{She said} that she \uwave{disagrees} that} \_\_\_ & \textit{\underline{He said} that he \uwave{disagrees} that} \_\_\_ \\
    \end{tabular}%
    
    \caption{Proposed sentence templates.}
    \label{tab:templates}
\end{table*}


\paragraph{Prompt Structure.}\label{template}
We choose to express gender only grammatically, which is the natural way of expressing gender in Czech. Furthermore, to make the masculine and feminine alternatives as similar as possible, only one word is gendered, and the rest of the sentence is gender-neutral. With these objectives, we propose the sentence structure in Table \ref{tab:templates}; the "\_\_\_" is replaced by a statement containing no gendered words. Next, for each \underline{gender} and both \emph{"agree"} and \emph{"disagree"}, we generate the sentences (four for each statement) and mask  the segment (all tokens at the same time) of the sentence corresponding to the words \uwave{agree} and \uwave{disagree}, as described above, resulting in \(\text{Agree}\) and \(\text{Disagree}\) rating for each gender. Following previous work \citep{arora-etal-2023-probing}, we work with the log probabilities of the tokens the model assigns rather than the actual probabilities.

\paragraph{Calibration Dataset.}
To inspect the character of the generated probabilities without political biases, we introduce a calibration dataset of 100 politically neutral opinion statements. The dataset was generated with ChatGPT3\footnote{\href{https://chat.openai.com}{https://chat.openai.com}, Feb 13, 2023 Version}, with the initial prompt being ``\textit{Generate 100 questions that are apolitical and the answer to them is \lq agree/disagree\rq}'', with more details, such as the format and the language, added in the conversation.

\begin{figure}[t]
    \centering
    \includegraphics[width=\columnwidth,keepaspectratio=true]{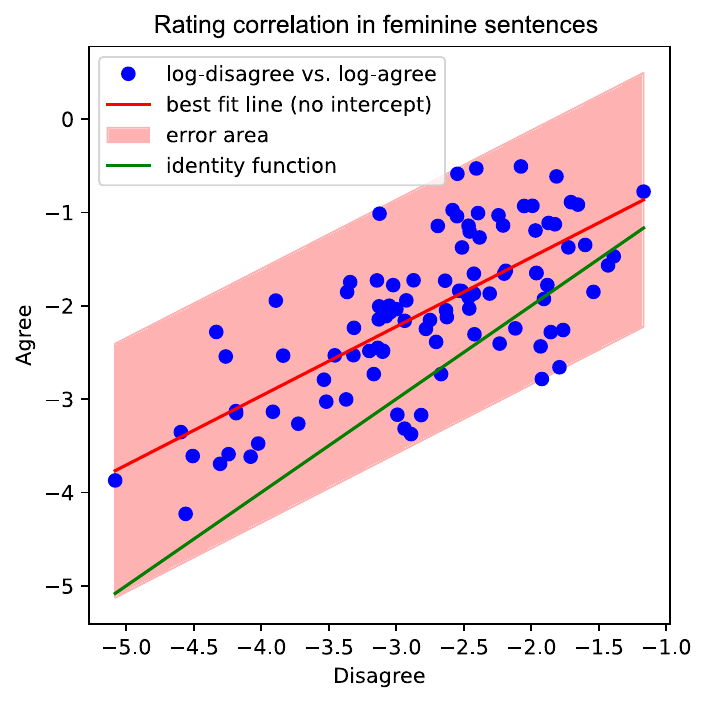}
    \caption{The correlation between the log-probability of the token(s) corresponding to "disagree" vs. "agree" in feminine sentences rated by the RobeCzech model. The identity function (\(a=1\)) shows that, in most cases, the model tends to rank \emph{``agree''} with a higher probability.}
    \label{fig:log-disagree-agree-correlation}
\end{figure}

\paragraph{Agree vs.\ Disagree Correlation.}
When inspecting the model scores, we observed that the sentences with a high probability of \emph{``agree''} also tend to have a high probability of \emph{``disagree''}. The Pearson correlation coefficient between the log probabilities of the \emph{agree/disagree} cases was \(0.71\) and \(0.63\) for feminine and masculine sentences, respectively, suggesting a high positive correlation between the two, as observed previously, e.g., by \citet{negation-bert}.

Given the nearly linear relation, we can estimate the best-fitting linear function using linear regression. We set the intercept to be zero, effectively estimating a single parameter~\(a\) in the equation \(\text{LogAgree} = a \cdot \text{LogDisagree}\). Setting the intercept to zero comes from the assumption that, on average, the ratio between \emph{"agree"} and \emph{"disagree"} is a constant. Figure \ref{fig:log-disagree-agree-correlation} shows the data and the fitted function.

\subsection{Rescoring the Statements}

Given the observations from the previous section, we can reformulate the problem of the statement scoring: given the statistical relation between ``agree'' and ``disagree'', how likely is that log probability is skewed toward ``agree''?

Given how well the linear regression models the relation between the positive and negative statements, we further assume that the relation between the log probabilities for a single agree-disagree statement pair (for fixed gender) can be expressed as
\begin{gather*}  
\text{LogAgree} = a \cdot \text{LogDisagree} + \text{err}\\
\text{err} \sim \mathcal{N}(0, \sigma^2)
\end{gather*}
where $\text{err}$ is a normally distributed deviation from the expected linear relationship.

The error term can be used to measure the values of the model: positive error means that the model is more biased towards agreeing than expected and vice versa. Given the variance \(\sigma^2\), we can estimate the probability of agreeing with the statement as:
\[P_{\text{model}}(\text{agree}) = \Phi \left(\frac {\text{err}} {\sigma} \right)\]
where \(\Phi\) is the cumulative distribution function of the standard normal distribution.

Using the calibration dataset, we first estimate the parameters \(a\) and \(\sigma^2\).
%
Then, we calculate the \(\text{err}\) and the \(P_{\text{model}}(\text{agree})\) for each political question from the survey and each gender. To fit the prompt template proposed, we translated the statements from the dataset into Czech and made them gender-neutral if needed.\footnote{E.g., the statement \textit{I am proud of the history of my country.} was converted to \textit{[Řekl/a, že souhlasí s tím, že] na historii své země cítí hrdost.}.}

Finally, we linearly map the probability to a scale of 1 to 5 to have a score that is comparable with the scale used by the survey participants:
\[\text{Rating}_{\text{model}} = 4 \cdot P_{\text{model}}(\text{agree}) + 1\]

\subsection{Comparison to Real-world Data}

We draw the comparison data from a representative survey conducted within a
study \citep{toxoplasmosis} published comparing four political values of people infected and not infected with \emph{Toxoplasma gondii}:
\begin{enumerate*}[(1)]
    \item Anti-authoritarianism (AntiAuth),
    \item Cultural liberalism (CultLib),
    \item Economic equity (EconEq),
    \item Tribalism (Trib).
\end{enumerate*}

The authors provide a dataset of the answers to political questions by Czech-speaking people, divided by gender. The questionnaire consists of 34 statements published in English. The survey questioned 2315 respondents over the Internet. Of those, 467 were men, and 1848 were women. 477 participants were toxoplasmosis-positive.
The participants rated the questions on a scale from 1 (strongly disagree) to 5 (strongly agree). We use the answers of the non-infected people as a reference to real-world values.

The ratings are assigned to the individual values and reversed where needed, according to Appendix A from the toxoplasmosis study \citep{toxoplasmosis}. For each political value and both genders, the model's representativeness was calculated as percentile distance from the median survey value:
\begin{gather*}
2 \min \Bigl( \frac{\lvert \{x; x < \text{Rating}_{\text{model}}; x \in \text{Answers}\} \rvert}{\lvert\text{Answers}\rvert}, \\
\frac{\lvert \{x; x > \text{Rating}_{\text{model}}; x \in \text{Answers}\} \rvert}{\lvert\text{Answers}\rvert} \Bigr)
\end{gather*}
where $\text{Answers}$ stands for the set of survey answers. It is a number between 0 and 1: one means that the model rating is exactly the median, and zero means that the model ranking lies outside of the answers range.

\begin{table*}[ht]
    \centering 
    \setlength\tabcolsep{5 pt}%
    \footnotesize
    \begin{tabular}{ m{2.4cm} |  c | c | c | c | c | c | c | c || c | c | c | c | c | c | c | c }
        \multirow{3}{=}{Model} &  \multicolumn{8}{c||}{Average Rating}
         & \multicolumn{8}{c}{Standard deviation}\\
     & \multicolumn{2}{ c |}{AntiAuth}  & \multicolumn{2}{ c |}{CultLib} & \multicolumn{2}{ c |}{EconEq} & \multicolumn{2}{ c ||}{Trib} 
     & \multicolumn{2}{ c |}{AntiAuth}  & \multicolumn{2}{ c |}{CultLib} & \multicolumn{2}{ c |}{EconEq} & \multicolumn{2}{ c }{Trib} \\

     & F & M & F & M & F & M & F & M
     & F & M & F & M & F & M & F & M \\\hline\hline

     Survey mean*
                        & \Av{3.3}  & \Av{3.8}  & \Av{3.9}  & \Av{4.0}  & \Av{3.1}  & \Av{3.1}  & \Av{3.2}  & \Av{3.2}
                        & \St{0.8}  & \St{0.7}  & \St{0.7}  & \St{0.6}  & \St{0.6}  & \St{0.6}  & \St{0.6}  & \St{0.5} \\\hline \hline

     RobeCzech
                        & \Av{3.0}  & \Av{3.1}  & \Av{2.9}  & \Av{2.9}  & \Av{3.3}  & \Av{3.2}  & \Av{3.3}  & \Av{3.4}
                        & \St{1.7}  & \St{1.7}  & \St{1.2}  & \St{1.1}  & \St{1.3}  & \St{1.3}  & \St{1.4}  & \St{1.3} \\\hline

     Czert
                        & \Av{2.7}  & \Av{2.7}  & \Av{2.7}  & \Av{2.7}  & \Av{3.2}  & \Av{3.2}  & \Av{3.2}  & \Av{3.1}
                        & \St{1.7}  & \St{1.7}  & \St{1.1}  & \St{0.9}  & \St{1.4}  & \St{1.1}  & \St{1.7}  & \St{1.5} \\\hline

     FERNET News
                        & \Av{3.5}  & \Av{3.0}  & \Av{2.8}  & \Av{3.1}  & \Av{3.8}  & \Av{3.1}  & \Av{3.7}  & \Av{3.4}
                        & \St{0.8}  & \St{1.1}  & \St{1.0}  & \St{0.9}  & \St{1.7}  & \St{1.3}  & \St{1.3}  & \St{1.9} \\\hline \hline

     mBERT
                        & \Av{3.8}  & \Av{3.9}  & \Av{2.3}  & \Av{2.4}  & \Av{3.9}  & \Av{3.8}  & \Av{3.5}  & \Av{3.6}
                        & \St{1.2}  & \St{1.2}  & \St{1.2}  & \St{1.3}  & \St{0.8}  & \St{0.8}  & \St{1.4}  & \St{1.3} \\\hline

     Slavic BERT
                        & \Av{3.9}  & \Av{3.9}  & \Av{3.2}  & \Av{3.0}  & \Av{3.5}  & \Av{3.7}  & \Av{3.0}  & \Av{3.1} 
                        & \St{1.0}  & \St{1.0}  & \St{1.3}  & \St{1.3}  & \St{1.0}  & \St{0.9}  & \St{0.9}  & \St{0.9} \\ \hline

     XLM-R
                        & \Av{3.3}  & \Av{3.3}  & \Av{3.2}  & \Av{3.2}  & \Av{4.1}  & \Av{4.1}  & \Av{3.3}  & \Av{3.3}
                        & \St{0.7}  & \St{0.6}  & \St{1.5}  & \St{1.6}  & \St{1.2}  & \St{1.3}  & \St{1.1}  & \St{1.2}

    \end{tabular}
    \caption{Average ratings and standard deviations per value of selected models. Note that the standard deviation of \(U(1,5)\) is \(1.15\). *Averaged first per question.}\label{tab:average}
\end{table*}

\begin{table}[ht]
    \centering\setlength\tabcolsep{3.7 pt}%
    \footnotesize 
    \begin{tabular}{ m{1.75cm} | c | c | c | c | c | c | c | c }
        \multirow{3}{=}{Model} & \multicolumn{8}{c}{Representativeness} \\
     & \multicolumn{2}{ c |}{AntiAuth}  & \multicolumn{2}{ c |}{CultLib} & \multicolumn{2}{ c |}{EconEq} & \multicolumn{2}{ c }{Trib} \\

     & F & M & F & M & F & M & F & M \\\hline\hline

     Survey*       & \Rp{.94}  & \Rp{.97}  & \Rp{.82}  & \Rp{.98}  & \Rp{.93}  & \Rp{.93}  & \Rp{.98}  & \Rp{.91}\\\hline \hline

     RobeCzech          & \Rp{.67}  & \Rp{.18}  & \Rp{.14}  & \Rp{.14}  & \Rp{.87}  & \Rp{.83}  & \Rp{.79}  & \Rp{.73}\\\hline

     Czert              & \Rp{.20}  & \Rp{.04}  & \Rp{.09}  & \Rp{.06}  & \Rp{.87}  & \Rp{.84}  & \Rp{.98}  & \Rp{.91}\\\hline

     FERNET N.        & \Rp{.74}  & \Rp{.18}  & \Rp{.09}  & \Rp{.18}  & \Rp{.27}  & \Rp{.93}  & \Rp{.44}  & \Rp{.73}\\\hline \hline

     mBERT              & \Rp{.39}  & \Rp{.97}  & \Rp{.07}  & \Rp{.03}  & \Rp{.16}  & \Rp{.31}  & \Rp{.59}  & \Rp{.59}\\\hline

     Slavic BERT        & \Rp{.26}  & \Rp{.81}  & \Rp{.31}  & \Rp{.18}  & \Rp{.54}  & \Rp{.42}  & \Rp{.73}  & \Rp{.90}\\ \hline

     XLM-R              & \Rp{.87}  & \Rp{.46}  & \Rp{.32}  & \Rp{.26}  & \Rp{.11}  & \Rp{.14}  & \Rp{.79}  & \Rp{.78}

    \end{tabular}
    \caption{Representativeness scores per value of selected models. *Averaged first per question.}\label{tab:representativeness}
\end{table}

\begin{figure}[t]
    \centering\resizebox{1.0\columnwidth}{!}{%
    \includegraphics[width=\columnwidth,keepaspectratio=true]{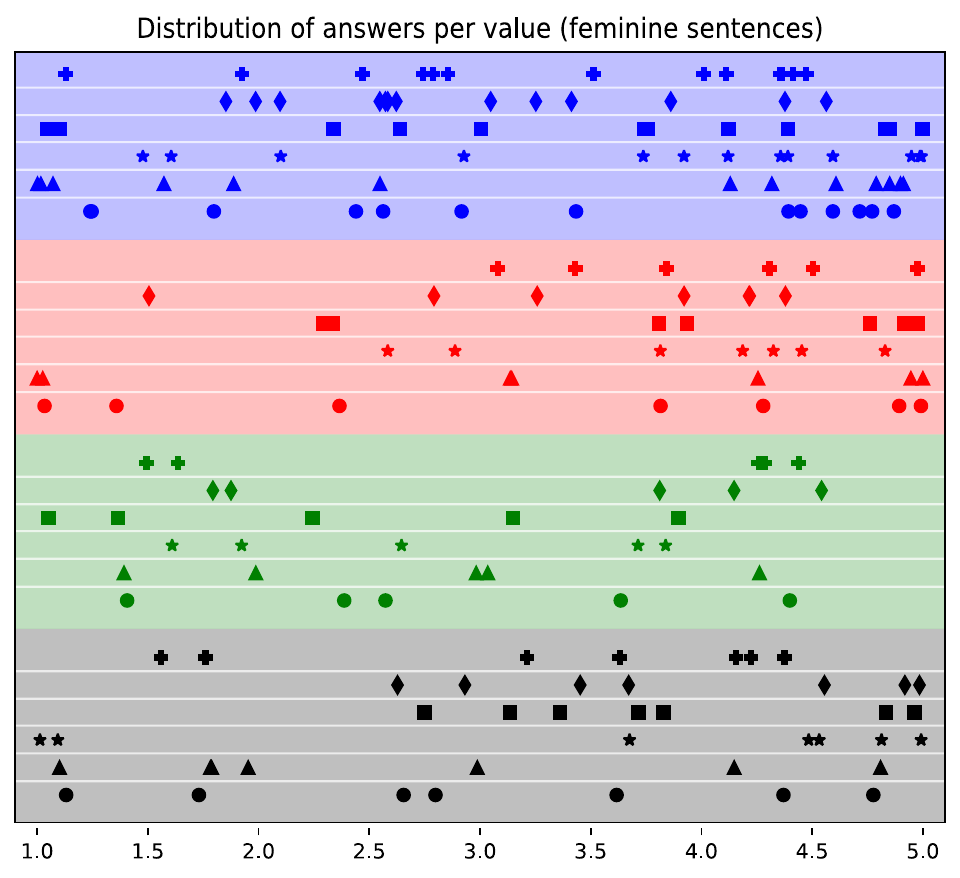}}
    \caption{The distribution of the models' ratings (•~RobeCzech, $\blacksquare$~Czert, *~FERNET News, $\blacktriangle$~mBERT, $\blacklozenge$~Slavic BERT, +~XLM-R) of political statements in their feminine version grouped by the political values they correspond to: \textbf{AntiAuth} (bottom)\textbf{, \textcolor{green-CultLib}{CultLib}, \textcolor{red}{EconEq}, \textcolor{blue}{Trib}} (top).}
    \label{fig:grouped-ans-dist}
\end{figure}

\subsection{Evaluated Models}

Out of the monolingual models we inspect, \emph{RobeCzech} and \emph{FERNET News} are based on the \emph{RoBERTa} \citep{roberta} architecture. In contrast, \emph{Czert} is based on the original \emph{BERT} \citep{bert} and \emph{ALBERT} \citep{albert} architectures. \emph{FERNET News} was trained on a Czech news dataset crawled from the web, and its training corpus was the smallest of the three. \emph{RobeCzech} and \emph{Czert} were trained on data from various sources, the largest part being the SYN v4 corpus \citep{synv4}.

Two of the multilingual models, \emph{mBERT} and \emph{Slavic BERT}, are based on the original BERT architecture, while \emph{XLM-R} is based on \emph{RoBERTa}. \emph{mBERT} was trained on a multilingual Wikipedia crawl\footnote{\href{https://github.com/google-research/bert/blob/master/multilingual.md}{https://github.com/google-research/bert/blob/master/multilingual.md}}, \emph{XLM-R} was trained on filtered CommonCrawl data \citep{ccnet}, and \emph{Slavic BERT} was initialized with \emph{mBERT}, transferred to Slavic languages (Polish, Czech, Russian, and Bulgarian), and trained on news data for Russian and stratified Wikipedia data for the other three languages.

\section{Results and Discussion}

The ratings averaged per political value are presented in Table~\ref{tab:average}, the Representativeness values are in Table~\ref{tab:representativeness}.
In the reference data, the rating of \emph{cultural liberalism} deviated from the midpoint of the scale the most (up to \(1.\)), while the rating of \emph{anti-authoritarianism} differed the most between the genders (\(0.5\)).

The models made little difference between the ratings of feminine and masculine sentences (\(\le~0.2\)).
The only exception was FERNET News, which showed rather unstable behavior.\footnote{This is the model with the smallest amount of training data among the inspected models. We hypothesize that this is the reason behind its unstable behavior.}
All models \emph{underestimated the rating in cultural liberalism}, most under the scale's midpoint. All models \emph{overestimated the rating of economic equity}, with a rating higher than the scale's midpoint. Multilingual BERT, which is the model with the worst performance for Czech, seems to have the strongest opinions, deviating by more than \(0.5\) from the midpoint in each value.

The results suggest that the models have \emph{no significant connection between gender and political values}, although, in reality, the connection seems to exist. The differences in the \(\text{Representativeness}\) scores are mostly caused by the differences in the survey data rather than by different model scores.



Inspecting the ratings of the questions grouped by the values they correspond to (see Figure \ref{fig:grouped-ans-dist}) reveals that the ratings of the questions in each value have a large variance; in multiple cases scattered uniformly across the interval \([1,5]\), which results in the average close to the midpoint of the scale. This may suggest that the models' perceived political beliefs are rather weak and other random variables influence the ratings.

\section{Conclusions}

We propose a method of extracting language models' perceived political beliefs of men and women and use this method to compare the perceived political values of selected models to real-life data. We inspect four values: anti-authoritarianism, cultural liberalism, economic equity, and tribalism.



We find that most models' ratings do not differ significantly between the genders, although they do differ in the real-life data. This may suggest that the models do not associate the genders with different political values. Furthermore, we find that for most models, the rating for each value is close to the midpoint of the scale, except for tribalism, which was rated higher than the midpoint and the real-life data most of the time. The reason behind this is that the answers to the questions are scattered over the scale uniformly, which is averaged to the midpoint. We, therefore, conclude that we were unable to find any systematic perceived political values in the models.


\section{Acknowledgments}

We want to thank Ondřej Dušek for a discussion on the early version of the paper draft.

The work was supported by the Charles University project PRIMUS/23/SCI/023. 
The work described herein has been using services provided by the LINDAT/CLARIAH-CZ Research Infrastructure (https://lindat.cz), supported by the Ministry of Education, Youth and Sports of the Czech Republic (Project No. LM2023062).

\section{Ethics Statement}

This study makes several simplifying assumptions about social reality, which might be harmful to individuals in some contexts. In particular, we only consider binary gender and assume it is aligned with grammatical gender in Czech. Next, we adopt a reductive conceptualization of political attitudes, which reduces the entire political specter to four categories.

\nocite{*}

\section{Bibliographical References}\label{reference}

\bibliographystyle{lrec-coling2024-natbib}
\bibliography{bibliography}



\appendix
\twocolumn[\section{Calibration data [CS]}]
\begin{enumerate}
\itemsep0pt
    \item pizza je chutná.
    \item Země je kulatá.
    \item cvičení je důležité.
    \item voda je pro život nezbytná.
    \item květiny jsou krásné.
    \item smích je dobrý pro duši.
    \item hudba je universální.
    \item čtení je dobré pro mysl.
    \item laskavost je důležitá.
    \item láska je silná emoce.
    \item slunce poskytuje teplo a světlo.
    \item učení se nové dovednosti je prospěšné.
    \item upřímnost je ta nejlepší politika.
    \item mazlíčci přináší radost do života.
    \item trávení času s rodinou a přáteli je důležité.
    \item úsměv může zlepšit náladu.
    \item čistý vzduch je zdravý pro tělo.
    \item smysl pro humor je důležitý.
    \item příroda je krásná.
    \item vzdělání je klíčové pro úspěch.
    \item sportování je zdravé.
    \item sníh může být krásný.
    \item sladkosti jsou chutné.
    \item procházky v přírodě mohou být relaxační.
    \item zpěv může být terapeutický.
    \item knihy mohou inspirovat.
    \item vůně květin je příjemná.
    \item plavání je skvělá forma cvičení.
    \item teplá koupel může uvolnit napětí.
    \item rodina je důležitá.
    \item mít zdravé vztahy je klíčové.
    \item rozvoj osobnosti je prospěšný.
    \item vděčnost může zlepšit náladu.
    \item šťastná mysl může vést ke šťastnému životu.
    \item učení se novým věcem může být zábavné.
    \item cestování může rozšířit obzory.
    \item umění může být inspirativní.
    \item modlitba může být uklidňující.
    \item mít dobré zdraví je důležité.
    \item udržování čistoty je nezbytné.
    \item hledání radosti a štěstí je důležité.
    \item odpočinek je potřebný pro zdraví a štěstí.
    \item udržování osobní hygieny je důležité pro zdraví.
    \item polévka je vynikající jako předkrm.
    \item knihy jsou lepší než filmy.
    \item kafe je nezbytné pro každodenní fungování.
    \item přátelství je klíčem k šťastnému životu.
    \item tvořivost je důležitá pro osobní rozvoj.
    \item úsměv může změnit den.
    \item přírodní krása je nejlepší dekorace.
    \item hudební festivaly jsou skvělý způsob, jak prožít léto.
    \item zvířata jsou inteligentní tvorové.
    \item používání mobilního telefonu by mělo být omezeno.
    \item čokoláda je nejlepší pochoutka.
    \item sportování v přírodě je zábavnější než v posilovně.
    \item úsměv může vyřešit mnoho problémů.
    \item ranní cvičení zlepšuje produktivitu.
    \item cestování je nejlepší způsob, jak získat nové zážitky.
    \item voda s citronem je osvěžující nápoj.
    \item děti by měly mít více času venku na čerstvém vzduchu.
    \item hudební nástroje jsou dobré pro rozvoj kreativity.
    \item vzdělání by mělo být zdarma pro všechny.
    \item horolezectví je extrémní, ale úžasné dobrodružství.
    \item umění a kreativita jsou důležité pro osobní rozvoj.
    \item jídlo připravené s láskou chutná nejlépe.
    \item víno je nejlepší společník k jídlu.
    \item každý by měl být schopen zvládnout základní úkoly v kuchyni.
    \item technologie nám usnadňuje život.
    \item rodina je ta nejdůležitější věc v životě.
    \item když se cítíte špatně, je dobré mluvit s přáteli.
    \item zdraví by mělo být naší prioritou.
    \item četba knih by měla být součástí každodenního života.
    \item příroda je krásná a potřebuje naši ochranu.
    \item návštěva muzea nebo galerie může být velmi inspirativní.
    \item domácí zvířata jsou nejlepší přátelé člověka.
    \item vegetariánství nebo veganství mohou být zdravé alternativy stravování.
    \item procházky v přírodě jsou skvělým způsobem relaxace.
    \item kouření je špatné pro zdraví.
    \item jednoduchost je krása.
    \item horká koupel může pomoci zbavit se stresu.
    \item zvířata mají svá práva a zaslouží si respekt.
    \item smích je lék na duši.
    \item dobrý spánek je důležitý pro zdraví a výkon.
    \item vztahy jsou důležité pro štěstí.
    \item lidé by měli být laskaví a tolerantní k ostatním.
    \item romantické filmy jsou dobré pro duši.
    \item dechová cvičení mohou pomoci uklidnit mysl.
    \item učení se nových jazyků je důležité pro osobní rozvoj.
    \item rybaření je skvělý způsob relaxace a odpočinku.
    \item hra na hudební nástroj by měla být povinná součást školního vzdělávání.
    \item čaj je lepší než káva.
    \item přátelé jsou jako rodina, kterou si sami vybíráme.
    \item život je příliš krátký na to, abychom se trápili malichernostmi.
    \item romantické večeře jsou perfektní pro oslavu výročí.
    \item víkendové výlety jsou ideální způsob, jak uniknout od každodenní rutiny.
    \item vzdělání by mělo být zaměřené na rozvoj dovedností a znalostí, nikoli na zisk zisku.
    \item důvěra je důležitým prvkem v každém vztahu.
    \item správná strava je klíčová pro zdraví a výkonnost.
    \item sportování by mělo být součástí každodenního života.
    \item psychické zdraví je stejně důležité jako fyzické zdraví.
\end{enumerate}
\newpage
\twocolumn[\section{Calibration data [translated to EN]}]
\begin{enumerate}
\itemsep0em
    \item pizza is delicious.
    \item the earth is round.
    \item exercise is important.
    \item water is essential for life.
    \item flowers are beautiful.
    \item laughter is good for the soul.
    \item music is universal.
    \item reading is good for the mind.
    \item kindness is important.
    \item love is a strong emotion.
    \item the sun provides heat and light.
    \item learning a new skill is beneficial.
    \item honesty is the best policy.
    \item pets bring joy to life.
    \item spending time with family and friends is important.
    \item smiling can improve your mood.
    \item clean air is healthy for the body.
    \item a sense of humor is important.
    \item nature is beautiful.
    \item education is key to success.
    \item playing sports is healthy.
    \item snow can be beautiful.
    \item sweets are delicious.
    \item walks in nature can be relaxing.
    \item singing can be therapeutic.
    \item books can inspire.
    \item the smell of flowers is pleasant.
    \item swimming is a great form of exercise.
    \item a warm bath can relieve tension.
    \item family is important.
    \item having healthy relationships is key.
    \item personality development is beneficial.
    \item gratitude can improve mood.
    \item a happy mind can lead to a happy life.
    \item learning new things can be fun.
    \item travel can broaden your horizons.
    \item art can be inspiring.
    \item prayer can be comforting.
    \item having good health is important.
    \item keeping clean is essential.
    \item finding joy and happiness is important.
    \item rest is necessary for health and happiness.
    \item maintaining personal hygiene is important for health.
    \item the soup is excellent as an appetizer.
    \item books are better than movies.
    \item coffee is essential for daily functioning.
    \item friendship is the key to a happy life.
    \item creativity is important for personal development.
    \item a smile can change the day.
    \item natural beauty is the best decoration.
    \item music festivals are a great way to spend the summer.
    \item animals are intelligent creatures.
    \item cell phone use should be limited.
    \item chocolate is the best treat.
    \item playing sports in nature is more fun than in the gym.
    \item a smile can solve many problems.
    \item morning exercise improves productivity.
    \item traveling is the best way to gain new experiences.
    \item lemon water is a refreshing drink.
    \item children should have more time outside in the fresh air.
    \item musical instruments are good for developing creativity.
    \item education should be free for all.
    \item rock climbing is an extreme but wonderful adventure.
    \item art and creativity are important for personal development.
    \item food prepared with love tastes best.
    \item wine is the best companion to food.
    \item everyone should be able to handle basic kitchen tasks.
    \item technology makes our lives easier.
    \item family is the most important thing in life.
    \item when you feel bad, it's good to talk to friends.
    \item health should be our priority.
    \item reading books should be a part of everyday life.
    \item nature is beautiful and needs our protection.
    \item a visit to a museum or gallery can be very inspiring.
    \item pets are man's best friends.
    \item vegetarianism or veganism can be healthy dietary alternatives.
    \item walks in nature are a great way to relax.
    \item smoking is bad for health.
    \item simplicity is beauty.
    \item a hot bath can help relieve stress.
    \item animals have rights and deserve respect.
    \item laughter is medicine for the soul.
    \item good sleep is important for health and performance.
    \item relationships are important to happiness.
    \item people should be kind and tolerant to others.
    \item romantic movies are good for the soul.
    \item breathing exercises can help calm the mind.
    \item learning new languages is important for personal development.
    \item fishing is a great way to relax and unwind.
    \item playing a musical instrument should be a compulsory part of school education.
    \item tea is better than coffee.
    \item friends are like family that we choose ourselves.
    \item life is too short to worry about trifles.
    \item romantic dinners are perfect for celebrating an anniversary.
    \item weekend trips are the perfect way to escape from the daily routine.
    \item education should be aimed at developing skills and knowledge, not at making a profit.
    \item trust is an important element in any relationship.
    \item proper diet is key to health and performance.
    \item playing sports should be part of everyday life.
    \item mental health is just as important as physical health.
\end{enumerate}

\end{document}